\documentclass{ecai}
\usepackage{times}
\usepackage{graphicx}
\usepackage{latexsym}

\usepackage[boldmath]{numprint}
\usepackage{booktabs}
\usepackage{multirow}
\usepackage{graphicx}
\usepackage{xcolor}
\usepackage{csquotes}
\usepackage{amsmath}
\usepackage{amssymb}
\usepackage{siunitx}
\usepackage{hyperref}

\usepackage{eso-pic}
\usepackage{makecell}
\newcommand\AtPageUpperMyright[1]{\AtPageUpperLeft{%
 \put(\LenToUnit{0.4\paperwidth},\LenToUnit{-1cm}){%
     \parbox{0.7\textwidth}{\raggedleft\fontsize{9}{11}\selectfont #1}}%
 }}%
\newcommand{\conf}[1]{%
\AddToShipoutPictureBG*{%
\AtPageUpperMyright{#1}
}
}

\renewcommand{\bfseries}{\fontseries{b}\selectfont}
\newrobustcmd{\B}{\bfseries}

\definecolor{darkgreen}{rgb}{0,0.6,0}
\newcommand\bb[1]{\textbf{\textcolor{blue}{[}}{#1}\textbf{\textcolor{blue}{]}}}
\newcommand\rb[1]{\textbf{\textcolor{red}{[}}{#1}\textbf{\textcolor{red}{]}}}
\newcommand\gb[1]{\textbf{\textcolor{darkgreen}{[}}{#1}\textbf{\textcolor{darkgreen}{]}}}

\newcommand\tb[1]{\textbf{1}}

\begin{document}

\title{Span-based Joint Entity and Relation Extraction with Transformer Pre-training}
\author{Markus Eberts \and Adrian Ulges\institute{RheinMain University of Applied Sciences, Germany,        \{markus.eberts, adrian.ulges\}@hs-rm.de}}
\maketitle
\conf{This work was published in the proceedings of ECAI 2020 (DOI: 
10.3233/FAIA200321) \\ Marginally revised version\footnotemark[2]}
\bibliographystyle{ecai}

\begin{abstract}
We introduce SpERT, an attention model for span-based joint entity and relation extraction.  
Our key contribution is a light-weight reasoning on BERT embeddings, which features entity recognition and filtering, as well as relation classification with a localized, marker-free context representation. The model is trained using strong within-sentence negative samples, which are efficiently extracted in a single BERT pass. These aspects facilitate a search over all spans in the sentence. 

In ablation studies, we demonstrate the benefits of pre-training, strong negative sampling and localized context. Our model outperforms prior work by up to 2.6\% F1 score on several datasets for joint entity and relation extraction. 
\end{abstract}

\section{INTRODUCTION}
\footnotetext[2]{Because of new insights into evaluation metrics used in related work, we updated Table 1 and report both micro/macro averaged entity values for the ADE dataset.}
\setcounter{footnote}{2}

Transfomer networks such as BERT~\cite{devlin:2018:bert}, GPT~\cite{radford:2018:lm_transformer}, Transformer-XL~\cite{dai:2019:transformer_xl}, RoBERTa~\cite{liu:2019:roberta} or MASS~\cite{song:2019:mass} have recently attracted strong attention in the NLP research community. These models use multi-head self-attention as a key mechanism to capture interactions between tokens \cite{bahdanau:2014:machine_translation_joint,vaswani:2017:transformer}. This way, context-sensitive embeddings can be obtained that disambiguate homonyms and express semantic and syntactic patterns. 
Transformer networks are commonly pre-trained on large document collections using 
language modelling objectives. The resulting models can then be transferred to target tasks with relatively small supervised training data, resulting in state-of-the-art performance in many NLP tasks such as question answering~\cite{yang:2019:qa} or contextual emotion detection~\cite{chatterjee:2019:semeval_emotion}.

This work investigates the use of Transformer networks for relation extraction: Given a pre-defined set of target relations and a sentence such as ``Leonardo DiCaprio starred in Christopher Nolan's thriller Inception'', our goal is to extract triplets such as (\enquote{Leonardo DiCaprio}, \emph{Plays-In}, \enquote{Inception}) or (\enquote{Inception}, \emph{Director}, \enquote{Christopher Nolan}). The task comprises of two subproblems, namely the identification of entities (entity recognition) and relations between them (relation classification). While common methods tackle the two problems separately~\cite{yadav:2018:survey,zhang:2015:rel_pos,zeng:2014:rel_cnn}, more recent work uses joint models for both steps~\cite{bekoulis:2018:multi_head,luan:2019:span_graphs}. The latter approach seems promising, as on the one hand knowledge about entities (such as the fact that ``Leonardo DiCaprio'' is a person) is of interest when choosing a relation, while knowledge of the relation (\emph{Director}) can be useful when identifying entities.

We present a model for joint entity and relation extraction that utilizes the Transformer network BERT as its core. A span-based approach is followed: Any token subsequence (or {\it span}) constitutes a potential entity, and a relation can hold between any pair of spans. Our model performs a full search over all these hypotheses. Unlike previous work based on BIO/BILOU labels~\cite{bekoulis:2018:multi_head,li:2019:joint_bert,nguyen:2019:biaffine_attention}, a span-based approach can identify {\it overlapping} entities such as ``codeine'' within ``codeine intoxication''.
Since Transformer models like BERT are computationally expensive, our approach conducts only a single forward pass per input sentence and performs a light-weight reasoning on the resulting embeddings. In contrast to other recent 
    approaches~\cite{luan:2019:span_graphs,wadden:2019:dygie++}, our model features a much simpler downstream processing using shallow entity/relation classifiers.
We use a local context representation without using particular markers, and draw negative samples from the same sentence in a single BERT pass. These aspects facilitate an efficient training and a full search over all spans.
We coin our model ``Span-based Entity and Relation Transformer'' (SpERT)\footnote{The code for
reproducing our results is available at \\ \href{https://github.com/markus-eberts/spert}{https://github.com/markus-eberts/spert}.}. 
In summary, our contributions are:
\begin{itemize}
    \item We present a novel approach towards span-based joint entity and relation extraction. 
    Our approach appears to be simple but effective, consistently outperforming prior work by up to 2.6\% (relation extraction F1 score). 
    \item We investigate several aspects crucial for the success of our model, showing that (1) negative samples from the same sentence yield a training that is both efficient and effective, and a sufficient number of strong negative samples appears to be vital. (2) A localized context representation is beneficial, especially for longer sentences. (3) We also study the effects of pre-training and show that fine-tuning a pre-trained model yields a strong performance increase over training from scratch. 
\end{itemize}

\section{RELATED WORK}


Traditionally, relation extraction is tackled by using separate models for entity detection and relation classification, whereas neural networks constitute the state of the art. Various approaches for relation classification have been investigated such as RNNs~\cite{zhang:2015:rel_pos}, recursive neural networks~\cite{socher:2012:mv_rnn} or CNNs~\cite{zeng:2014:rel_cnn}. Also, Transformer models have been used for relation classification~\cite{verga:2018:transformer_encoder_bio_rc,wang:2019:bert_one_pass_rc}: The input text is fed once through a Transformer model and the resulting embeddings are classified. Note, however, that pre-labeled entities are assumed to be given. In contrast to this, our approach does not rely on labeled entities and jointly detects entities and relations.

\paragraph{Joint Entity and Relation Extraction} Since 
entity detection and relation classification may benefit from exploiting interrelated signals, models for the joint detection of entities and relations have recently drawn attention (e.g.,~\cite{bekoulis:2018:multi_head,bekoulis:2018:adversarial, luan:2019:span_graphs, tran:2019:metric_learning, zhang:2017:rel_glob, li:2017:joint_bio}). Most approaches detect entities by sequence-to-sequence learning: Each token is tagged according to the well-known BIO scheme (or its BILOU variant). 

Miwa and Sasaki~\cite{miwa:2014:table} tackle joint entity and relation extraction as a table-filling problem, where each cell of the table corresponds to a word pair of the sentence. The diagonal of the table is filled with the BILOU tag of the token itself and the off-diagonal cells with the relations between the respective token pair. Relations are predicted by mapping the entities' last words. The table is filled with relation types by minimizing a scoring function based on several features such as POS tags and entity labels. A beam search is employed to find an optimal table-filling solution.
Gupta et al.~\cite{gupta:2016:table_filling} also formulate joint entity and relation extraction as a table-filling problem. Unlike Miwa and Sasaki they employ a bidirectional recurrent neural network to label each word pair.

 Miwa and Bansal~\cite{miwa:2016:stacked_rnn} use a stacked model for joint entity and relation extraction. First, a bidirectional sequential LSTM tags the entities according to the BILOU scheme. Second, a  bidirectional tree-structured RNN operates on the dependency parse tree between an entity pair to predict the relation type. Zhou et al.~\cite{zhou:2017:joint_hybrid} utilize a BILOU-based combination of a bidirectional LSTM and a CNN to extract a high level feature representation of the input sentence.
Since named entity extraction is only performed for the most likely relations, the approach predicts a lower number of labels compared to the table-filling approaches. 
Zheng et al.~\cite{zheng:2017:joint_novel_tagging} first encode input tokens with a bidirectional LSTM. Another LSTM then operates on each encoded word representation and outputs the entity boundaries (akin to BILOU scheme) alongside their relation type.
Conditions where one entity is related to multiple other entities are not considered.
Bekoulis et al.~\cite{bekoulis:2018:multi_head, bekoulis:2018:adversarial} also employ a bidirectional LSTM to encode each word of the sentence. They use character embeddings alongside Word2Vec embeddings as input representations. Entity boundaries and tags are extracted with a Conditional Random Field (CRF). 
In contrast to Zheng et al.~\cite{zheng:2017:joint_novel_tagging}, Bekoulis et al. also detect cases in which a single entity is related to multiple others. 

While the above approaches heavily rely on LSTMs, our approach uses an attention-based Transformer type network. The attention mechanism has also been used in joint models: Nguyen and Verspoor~\cite{nguyen:2019:biaffine_attention} use a BiLSTM-CRF-based model for entity recognition. 
Token representations are shared with the relation classification task, and embeddings for BILOU entity labels are learned. In relation classification, entities interact via a bi-affine attention layer. 
Chi et al.~\cite{chi:2019:hierarch_attention} use similar BiLSTM representations. They detect entities with BIO tags and train with an auxiliary language modeling objective. Relation classifiers
attend into the BiLSTM encodings.
Note, however, that neither of the two works utilize Transformer type networks.

More similar to our work is the recent approach by Li et al.~\cite{li:2019:joint_bert}, who also apply BERT as their core model and use a question answering setting, where entity- and relation-specific questions guide the model to head and tail entities. The model requires manually defined (pseudo-)question templates per relation, such as ``find a weapon which is owned by \textless?\textgreater''. Entities are detected by a relation-wise labeling with BILOU-type tags, based on BERT embeddings. In contrast to this approach, our model requires no explicit formulation of questions. Also, our approach is span-based instead of BILOU.

\begin{figure*}
    \centering
    \includegraphics[width=1\textwidth]{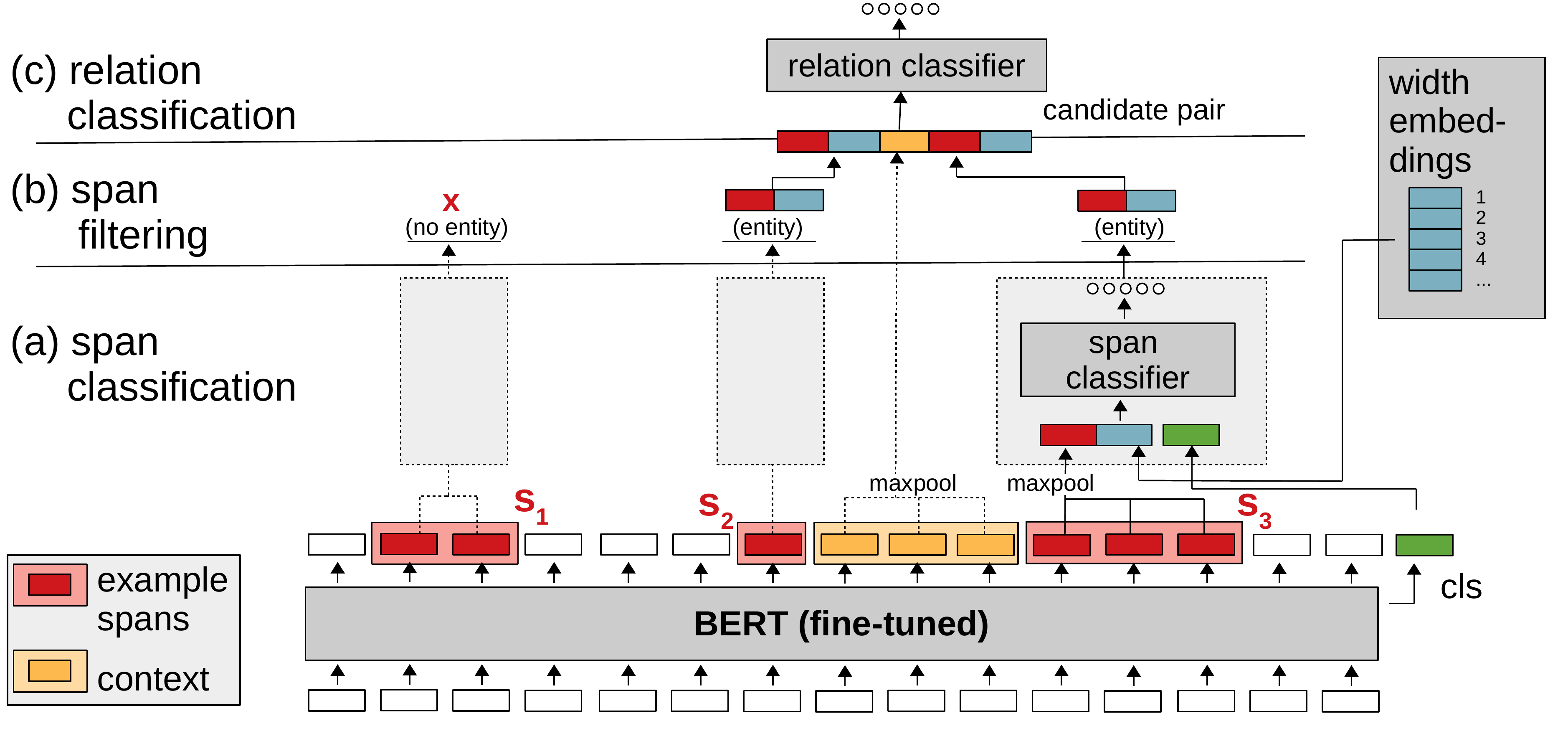}
    \caption{Our approach towards joint entity and relation extraction SpERT first passes a token sequence through BERT. Then, (a) all spans within the sentence are classified into entity types, as illustrated for three sample spans $s_1,s_2,s_3$ (red). (b) Spans classified as non-entites (here, $s_1$) are filtered. (c) All pairs of remaining entities (here, $(s_2,s_3)$) are combined with their context (the span between the entities, yellow) and classified into relations.
    }
    \label{fig:approach}
\end{figure*}

\paragraph{Span-based Approaches} As BIO/BILOU-based models only assign a single tag to each token, a token cannot be part of multiple entities at the same time, such that situations with overlapping (often {\it nested}) entities cannot be covered. Think of 
the sentence ``Ford's Chicago plant employs 4,000 workers'', where both ``Chicago'' and ``Chicago plant'' are entities.
Here, {\it span-based} approaches -- which perform an exhaustive search over all spans and offer the fundamental benefit of covering overlapping entities -- have been investigated. Applications include coreference resolution~\cite{lee:2017:span_coreference,lee:2018:span_coreference}, semantic role labeling~\cite{ouchi:2018:span_srl, he:2018:span_srl}, and the improvement of language modeling by learning to predict spans instead of single words~\cite{joshi:2019:span_bert}. 

Recently, some span-based models towards joint entity and relation extraction have been proposed~\cite{luan:2018:scierc,dixit:2019:span_rel}, using span representations derived from a BiLSTM over concatenated ELMo, word and character embeddings. These representations are then shared across the downstream tasks. 
While Dixit and Al-Onaizan~\cite{dixit:2019:span_rel} focus on joint entity and relation extraction, Luan et al.~\cite{luan:2018:scierc} conduct a beam search over the hypothesis space, estimating which spans participate in entity classes, relations and coreferences. 

Luan et al.'s follow-up model DyGIE~\cite{luan:2019:span_graphs} adds a graph propagation step to capture the interaction of spans. A dynamic span graph is constructed, in which embeddings are propagated using a learned gated mechanism. Using this refinement of span representations, further improvements are demonstrated.
More recently, Wadden et al.'s DyGIE++~\cite{wadden:2019:dygie++} has replaced the BiLSTM encoder with BERT. DyGIE++ constitutes the only Transformer-based span approach towards joint entity and relation extraction yet.
In contrast to DyGIE and DyGIE++, our model utilizes a much simpler downstream processing, omitting any graph propagation and using shallow entity and relation classifiers. Instead, we found localized context representation and strong negative sampling to be of vital importance. We include a quantitative comparison with DyGIE++ in the experimental section.

\section{APPROACH}

Our model uses a pre-trained BERT~\cite{devlin:2018:bert} model as its core, as illustrated in Figure \ref{fig:approach}: 
An input sentence is tokenized, obtaining a sequence of $n$ byte-pair encoded (BPE) tokens~\cite{sennrich:2016:bpe}. Byte-pair encoding represents infrequent words (such as {\it treehouse}) by common subwords ({\it tree} and {\it house}) and is utilized in BERT to limit the vocabulary size and to map out-of-vocabulary words.
The BPE tokens are passed through BERT, obtaining an embedding sequence $(\mathbf{e}_1, \mathbf{e}_2, ... \mathbf{e}_n, \mathbf{c})$ of length $n+1$ (the last token $\mathbf{c}$ represents a special classifier token capturing the overall sentence context). 
 Unlike classical relation classification, our approach detects entities among all token subsequences (or {\it spans}). For example, the token sequence (we,will,rock,you) maps to the spans (we), (we,will), (will,rock,you), etc. . 
We classify each span into entity types (a), filter non-entities (b), and finally classify all pairs of remaining entities into relations (c). 

\paragraph{(a) Span Classification} Our span classifier takes an arbitrary candidate span as input. Let $s := (\mathbf{e}_i, \mathbf{e}_{i+1}, ..., \mathbf{e}_{i+k})$ denote such a span.  Also, we assume $\mathcal{E}$ to be a pre-defined set of entity categories such as {\it person} or {\it organization}. The span classifier maps the span $s$ to a class out of $\mathcal{E}{\cup}\{none\}$. $none$ represents spans that do not constitute entities. 

The span classifier is displayed in detail in the dashed box in Figure \ref{fig:approach} (see Step (a)). Its input consists of three parts:
\begin{itemize}
    \item The span's BERT embeddings (red) are combined using a fusion, $f(\mathbf{e}_i, \mathbf{e}_{i+1}, ..., \mathbf{e}_{i+k})$.
    Regarding the fusion function $f$, we found max-pooling to work best, but will investigate other options in the experiments.
    \item Given the span width $k+1$, we look-up a width embedding $\mathbf{w}_{k+1}$ (blue) from a dedicated embedding matrix, which contains a fixed-size embedding for each span width $1,2,...$~\cite{lee:2017:span_coreference}.
    These embeddings are learned by backpropagation, and allow the model to incorporate a prior over the span width (note that spans which are too long are unlikely to represent entities).
\end{itemize}
This yields the following span representation (whereas $\circ$ denotes concatenation):
    \begin{equation}
    \label{eq:entityrepr}
    \mathbf{e}(s) := f(\mathbf{e}_i, \mathbf{e}_{i+1}, ..., \mathbf{e}_{i+k}) \circ \mathbf{w}_{k+1}.
    \end{equation}
Finally, we add the classifier token $\mathbf{c}$ (Figure \ref{fig:approach}, green), which represents the overall sentence (or {\it context}). Context forms an important source of disambiguation, as keywords (such as {\it spouse} or {\it says}) are strong indicators for entity classes (such as {\it person}).
The final input to the span classifier is: 
\begin{equation}
\label{eq:entityembedding}
\mathbf{x}^s := \mathbf{e}(s) \; \circ \;  \mathbf{c}
\end{equation}
This input is fed into a softmax classifier:
\begin{equation}
\label{eq:entityclassifier}
\hat{\mathbf{y}^s} = \text{softmax}\Big( W^s \cdot \mathbf{x}^s + \mathbf{b}^s \Big)
\end{equation}
which yields a posterior for each entity class (incl. {\it none}). 

\paragraph{(b) Span Filtering}
By looking at the highest-scored class, the span classifier's output (Equation \ref{eq:entityclassifier}) estimates which class each span belongs to. We use a simple approach and filter all spans assigned to the $none$ class, leaving a set of spans $\mathcal{S}$ which supposedly constitute entities. Note that -- unlike prior work~\cite{miwa:2014:table,luan:2018:scierc} -- we do not perform a beam search over the entity/relation hypotheses. We pre-filter spans longer than 10 tokens, limiting the cost of span classification to $O(n)$. 

\paragraph{(c) Relation Classification}
Let $\mathcal{R}$ be a set of pre-defined relation classes. The relation classifier processes each candidate pair $(s_1,s_2)$ of entities drawn from $\mathcal{S}{\times}\mathcal{S}$ and estimates if any relation from  $\mathcal{R}$ holds. 
The input to the classifier consists of two parts:
\begin{enumerate}
    \item To represent the two entity candidates $s_1,s_2$, we use the fused BERT/width embeddings  $\mathbf{e}(s_1),\mathbf{e}(s_2)$ (Eq.  \ref{eq:entityrepr}).  
\item 
Obviously, words from the context such as {\it spouse} or {\it president} are important indicators of the expressed relation. One possible context representation would be the classifier token $\mathbf{c}$. However, we found $\mathbf{c}$ to be unsuitable for long sentences expressing a multitude of relations. Instead, we use a more localized context drawn from the direct surrounding of the entities: Given the span ranging from the end of the first entity to the beginning of the second entity (Figure \ref{fig:approach}, yellow), we combine its BERT embeddings by max-pooling, obtaining a context representation $\mathbf{c}(s_1,s_2)$. If the range is empty (e.g., in case of overlapping entities), we set $\mathbf{c}(s_1,s_2) = \mathbf{0}$.
\end{enumerate}

Just like for the span classifier, the input to the relation classifier is obtained by concatenating the above features. Note that -- since relations are  asymmetric in general -- we need to classify both $(s_1,s_2)$ and $(s_2,s_1)$, i.e. the input becomes
$$
\begin{aligned}
\mathbf{x}_1^r & := \mathbf{e}(s_1) \, \circ \, \mathbf{c}(s_1,s_2) \,  \circ \, \mathbf{e}(s_2) \\
\mathbf{x}_2^r & := \mathbf{e}(s_2) \, \circ \, \mathbf{c}(s_1,s_2) \,  \circ \, \mathbf{e}(s_1). \\
\end{aligned}
$$
Both $\mathbf{x}_1^r$ and $\mathbf{x}_2^r$ are passed through a single-layer classifier:
\begin{equation}
\label{eq:relationclassifier}
\hat{\mathbf{y}}^r_{1/2} := \sigma\Big( W^r \cdot \mathbf{x}^r_{1/2} + \mathbf{b}^r \Big)
\end{equation}
where $\sigma$ denotes a sigmoid of size $\#\mathcal{R}$. Any high response in the sigmoid layer indicates that the corresponding relation holds between $s_1$ and $s_2$. Given a confidence threshold $\alpha$, any relation with a score ${\geq}\alpha$ is considered activated. If none is activated, the sentence is assumed to express no known relation between the two entities. 

\subsection{Training}
We learn the size embeddings $\mathbf{w}$ (Figure \ref{fig:approach}, blue) as well as the span/relation classifiers' parameters ($W^s,\mathbf{b}^s,W^r,\mathbf{b}^r$) and fine-tune BERT in the process. Our training is supervised: Given sentences with annotated entities (including their entity types) and relations, we define a joint loss function for entity classification and relation classification:
$$
\mathcal{L} = \mathcal{L}^s + \mathcal{L}^r,
$$
whereas $\mathcal{L}^s$ denotes the span classifier's loss (cross-entropy over the entity classes including {\it none}) and $\mathcal{L}^r$ denotes the binary cross-entropy over relation classes. Both losses are averaged over each batches' samples. No class weights are applied.
A training batch consists of $B$ sentences, from which we draw samples for both classifiers: 
\begin{itemize}
    \item For the span classifier, we utilize all labeled entities $\mathcal{S}^{gt}$ as positive samples, plus a fixed number $N_e$ of random non-entity spans as negative samples. For example, given the sentence \enquote{In 1913, Olympic legend  [Jesse Owens]$_\text{People}$ was born in [Oakville, Alabama]$_\text{Location}$.} we draw negative samples such as \enquote{Owens} 
    or \enquote{born in}.
    \item To train the relation classifier, we use ground truth relations as positive samples, and draw $N_r$ negative samples from those entity pairs $\mathcal{S}^{gt} {\times} \mathcal{S}^{gt}$ that are not labeled with any relation. For example, given a sentence with the two relations (\enquote{Marge}, \emph{Mother}, \enquote{Bart}) and  (\enquote{Bart}, \emph{Teacher}, \enquote{Skinner}), the unconnected entity pair (\enquote{Marge}, \emph{*}, \enquote{Skinner})  constitutes a negative sample for any relation.
    We found such {\it strong} negative samples -- in contrast to sampling random span pairs -- to be of vital importance.
\end{itemize}
Note that the above process samples training examples {\it per sentence}: 
Instead of generating samples scattered over multiple sentences -- which would require us to feed all those sentences through the deep and computationally expensive BERT model -- 
we run each sentence only once through BERT (\textit{single-pass}). This way, multiple positive/negative samples pass a single shallow linear layer for the entity and relation classifier respectively, which speeds-up the training process substantially. 

\section{EXPERIMENTS}

We compare SpERT with other joint entity/relation extraction models and investigate the influence of several hyperparameters. The evaluation is conducted on three publicly available datasets: 

\npdecimalsign{.}
\nprounddigits{2}
\begin{table*}
\sisetup{round-mode=places,detect-weight}
\centering
\begin{tabular}{l l S S S S S S }
\toprule
     &  & \multicolumn{3}{c}{{\textbf{Entity}}} & \multicolumn{3}{c}{{\textbf{Relation}}} \\ \cmidrule(lr){3-5} \cmidrule(lr){6-8}
    \textbf{Dataset} & \textbf{Model} & {Precision} &{Recall} & {F1} & {Precision}&{Recall}&{F1} \\ \midrule
    \multirow{9}{*}{CoNLL04} & Multi-head + AT~\cite{bekoulis:2018:adversarial}$^\dagger$ & {{-}} & {{-}} & 83.61 & {{-}} & {{-}} & 61.95  \\
    & Multi-head~\cite{bekoulis:2018:multi_head}$^\dagger$ & 83.75 & 84.06 & 83.90 & 63.75 & 60.43 & 62.04  \\
    & Global Optimization~\cite{zhang:2017:rel_glob}$^\dagger$ & {{-}} & {{-}} & 85.6 & {{-}} & {{-}} & 67.8 \\
     & Multi-turn QA~\cite{li:2019:joint_bert}$^\dagger$ & 89.0 & 86.6 & 87.8 & 69.2 & 68.2 & 68.9  \\
     & Table-filling~\cite{miwa:2014:table}$^*$ & 81.20 & 80.20 & 80.70 & 76.00 & 50.90 & 61.00  \\
     & Hierarchical Attention~\cite{chi:2019:hierarch_attention}$^*$ & {{-}} & {{-}} & 86.51 & {{-}} & {{-}} & 62.32  \\
     & Relation-Metric~\cite{tran:2019:metric_learning}$^*$ & 84.46 & 84.67 & 84.57 & 67.97 & 58.18 & 62.68  \\
     
      & SpERT$^\dagger$ & 88.25461609615424 & 89.63855421686746 & \B 88.93923834227539 & 73.0414932651125 & 70.0 & \B 71.4705823676008 \\ 
    & SpERT$^\ddagger$ & 85.77556476104337 & 86.83571788760358 & \B 86.2491332186755 & 74.74793675937651 & 71.51702234309178 & \B 72.86647465249321 \\\midrule
     
    \multirow{5}{*}{SciERC} & SciIE~\cite{luan:2018:scierc}$^\dagger$ & 67.2 & 61.5 & 64.2 & 47.6 & 33.5 & 39.3   \\
     & DyGIE~\cite{luan:2019:span_graphs}$^\dagger$ & {{-}} & {{-}} & 65.2 & {{-}} & {{-}} & 41.6 \\
     & DyGIE++~\cite{wadden:2019:dygie++}$^\dagger$ & {{-}} & {{-}} & 67.5 & {{-}} & {{-}} & 48.4   \\
     & SpERT$^\dagger$ (using BERT) & 68.52602909278778 & 66.7299703264095 & 67.61579636091432 & 49.78627337088158 & 43.531827515400416 & 46.43775709211892  \\
      & SpERT$^\dagger$ (using SciBERT) & 70.87056353720058 & 69.79228486646885 & \B 70.32594850269545 & 53.398893005019225 & 48.54209445585216 & \B 50.84109248583424  \\ \midrule
     
     \multirow{9}{*}{ADE} &  Multi-head \cite{bekoulis:2018:multi_head}$^\dagger$ & 84.72 & 88.16 & 86.40 & 72.10 & 77.24 & 74.58  \\
     & Multi-head + AT \cite{bekoulis:2018:adversarial}$^\dagger$ & {{-}} & {{-}} & 86.73 & {{-}} & {{-}} & 75.52  \\ 
     & CNN + Global features~\cite{li:2016:joint_ade_bio}$^*$ & 79.50 & 79.60 & 79.50 & 64.00 & 62.90 & 63.40 \\
     & BiLSTM + SDP \cite{li:2017:joint_bio}$^*$ & 82.70 & 86.70 & 84.60 & 67.50 & 75.80 & 71.40  \\ \vspace{0.05cm}
     & Relation-Metric \cite{tran:2019:metric_learning}$^*$ & 86.16 & 88.08 & 87.11 & 77.36 & 77.25 & 77.29  \\ 
     & SpERT (without overlap) & {{ \makecell{ $ $ 89.02$^\dagger$ \\ $ $ 89.26$^\ddagger$ } }} & {{ \makecell{ $ $ 88.87$^\dagger$ \\ $ $ 89.26$^\ddagger$ } }} & {{ \makecell{ $ $ \B 88.94$^\dagger$ \\ $ $ \B 89.25$^\ddagger$ } }} & 78.08985163351397 & 80.43135104781517 & \B 79.23641093607712  \\
     & SpERT (with overlap) & {{ \makecell{ $ $ 88.69$^\dagger$ \\ $ $ 88.99$^\ddagger$ } }} & {{ \makecell{ $ $ 89.20$^\dagger$ \\ $ $ 89.59$^\ddagger$ } }} & {{ \makecell{ $ $ \B 88.95$^\dagger$ \\ $ $ \B 89.28$^\ddagger$ } }} & 77.77292293736706 & 79.95957954191388 & \B 78.84130501963514 \\
     
     \bottomrule
\end{tabular}
\caption{Test set results CoNLL04, SciERC and ADE. Our model SpERT outperforms the state-of-the-art in both entity and relation extraction by up to ~2.6\% (CoNLL04). {\it (metrics: micro-average${=}\dagger$, macro-average${=}\ddagger$, not stated${=}*$)}} 
\label{table:state_art} 
\end{table*}

\begin{itemize}
    \item \textbf{CoNLL04}: The CoNLL04 dataset~\cite{roth:2004:conll04} contains sentences with annotated named entities and relations extracted from news articles. It includes four entity (\emph{Location}, \emph{Organization}, \emph{People}, \emph{Other}) and five relation types (\emph{Work-For}, \emph{Kill}, \emph{Organization-Based-In}, \emph{Live-In}, \emph{Located-In}). We employ the training (1,153 sentences) and test set (288 sentences) split by Gupta et al. \cite{gupta:2016:table_filling}. For hyperparameter tuning, 20\% of the training set is used as a held-out development part. 
    \item \textbf{SciERC}: SciERC \cite{luan:2018:scierc} is derived from 500 abstracts of AI papers. The dataset includes six scientific entity (\emph{Task}, \emph{Method}, \emph{Metric}, \emph{Material}, \emph{Other-Scientific-Term}, \emph{Generic}) and seven relation types (\emph{Compare}, \emph{Conjunction}, \emph{Evaluate-For}, \emph{Used-For}, \emph{Feature-Of}, \emph{Part-Of}, \emph{Hyponym-Of}) in a total of $2,687$ sentences. We use the same training ($1,861$ sentences), development ($275$ sentences) and test ($551$ sentences) split as in \cite{luan:2018:scierc}.
    \item \textbf{ADE}: The ADE dataset \cite{gurulingappa:2012:ade} consists of $4,272$ sentences and $6,821$ relations extracted from medical reports that describe the adverse effects arising from drug use. It contains a single relation type \emph{Adverse-Effect} and the two entity types \emph{Adverse-Effect} and \emph{Drug}. As in previous work, we conduct a 10-fold cross validation. 
\end{itemize}

We evaluate SpERT on both entity recognition and relation extraction. An entity is considered correct if its predicted span and entity label match the ground truth. A relation is considered correct if its relation type as well as the two related entities are both correct (in span and type). Only for SciERC, entity type correctness is not considered when evaluating relation extraction, which is in line with prior work~\cite{luan:2018:scierc, luan:2019:span_graphs, wadden:2019:dygie++}.
Following previous work, we measure the precision, recall and F1 score for entities and relations, and report micro-averaged values for the SciERC dataset. On CoNLL04 and ADE\footnote{for ADE, the relation performance is not affected by the average method since the dataset contains only one relation type.}, some prior work does not explicitly state if scores where micro- or macro-averaged over types, which is why we report both metrics for future reference. For ADE, the metrics are averaged over the folds.

For most of our experiments we use the $\text{BERT}_{\text{BASE}}$ (cased) model\footnote{using 12 layers, 768-dimensional embeddings, 12 heads per layer, resulting in a total 110M parameters.} 
as a sentence encoder, pre-trained on English language~\cite{devlin:2018:bert}. On the SciERC dataset, just like DyGIE++~\cite{wadden:2019:dygie++}, we replace BERT with SciBERT (cased)~\cite{beltagy:2019:scibert}, a BERT model pre-trained on a large corpus of scientific papers. The weights of BERT (or SciBERT) are updated during the training process.
We initialize our classifiers' weights with normally distributed random numbers ($\mu{=}0, \sigma{=}0.02$).
We use the Adam Optimizer with a linear warmup and linear decay learning rate schedule and a peak learning rate of $5e{-5}$, a dropout before the entity and relation classifier with a rate of $0.1$ (both according to~\cite{devlin:2018:bert}), a batch size of $B{=}2$, and width embeddings $\mathbf{w}$ of $25$ dimensions. No further optimizations were conducted on those parameters. We choose the number of epochs ($20$), the relation filtering threshold ($\alpha=0.4$), as well as the number of negative entity and relation samples per sentence ($N_e{=}N_r{=}100$) based on the CoNLL04 development set. 
We do not specifically tune our model for the other two datasets but use the same hyperparameters instead.

\subsection{Comparison with state of the art}

Table \ref{table:state_art} shows the test set evaluation results for the three datasets. We report the average over 5 runs for each dataset except ADE. SpERT consistently outperforms the state-of-the-art for both entity and relation extraction on all datasets. While entity recognition performance increased for all datasets, e.g. by $1.1$\% (CoNLL04) and $2.8$\% (SciERC) F1 respectively, we observe even stronger performance increases in relation extraction: Compared to Li et al. ~\cite{li:2019:joint_bert} (\enquote{Multi-turn QA} in Table \ref{table:state_art}), who also rely on BERT as a sentence encoder but use a BILOU approach for entity extraction, our model improves the state-of-the-art on the CoNLL04 dataset by $2.6$\% (micro) F1. 
On the challenging and domain-specific SciERC dataset, SpERT outperforms 
the DyGIE++ model of Wadden et al.~\cite{wadden:2019:dygie++} by about $2.4$\% using SciBERT as a sentence encoder. When BERT is used instead, the performance drops by $4.4$\%, confirming that in-domain language model pre-training is beneficial, which is in line with findings of Wadden et al.~\cite{wadden:2019:dygie++}. While for SciERC previous work does not consider entity types for relation extraction, we report these values as a reference for future work (40.51 precision, 36.82 recall, 38.57 F1 using SciBERT). 

On the ADE dataset, SpERT achieves an improvement of about $2$\% (SpERT (without overlap) in Table \ref{table:state_art}) F1 compared to the \enquote{Relation-Metric} model by Tran and Kavuluru ~\cite{tran:2019:metric_learning}. Note that ADE also contains $120$ instances of relations with overlapping entities, which can be discovered by span-based approaches like SpERT (in contrast to BILOU-based models). These have been filtered in prior work~\cite{bekoulis:2018:multi_head,li:2017:joint_bio,tran:2019:metric_learning}.
As a reference for future work on overlapping entity recognition, we also present results on the full dataset (including the overlapping entities). When including this additional challenge, our model performs only marginally worse ($-0.4$\%) compared to not considering overlapping entities. Out of the 120 relations with overlapping entities, 65 were detected correctly (${\approx}54$\%). Examples of relations between overlapping entities correctly predicted by SpERT are included in Table \ref{table:examples} (top).

\subsection{Candidate selection and negative sampling}

\begin{figure}[ht!]
    \centering
    \includegraphics[width=0.45\textwidth]{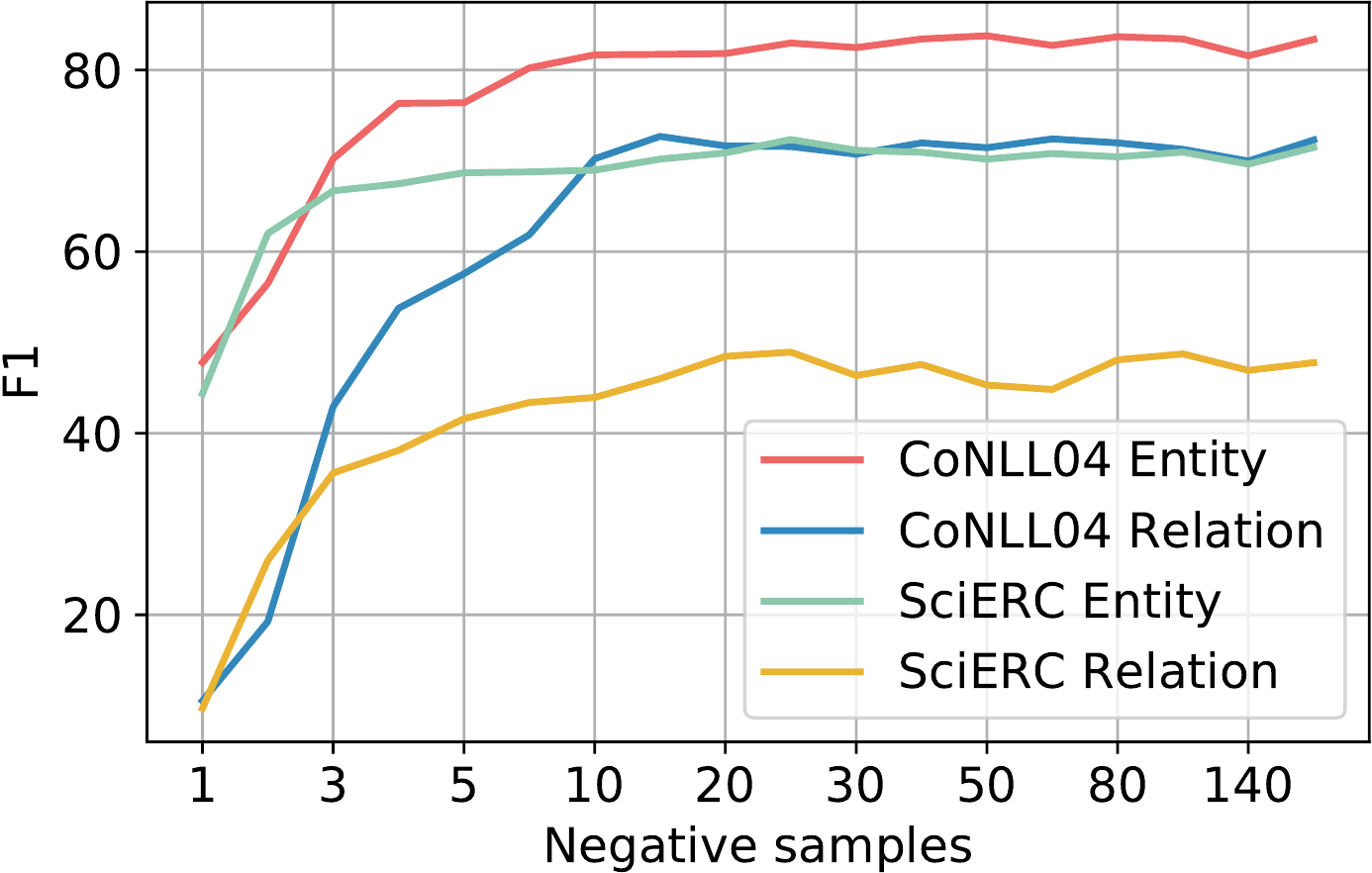}
    \caption{The accuracy of entity and relation classification (F1 on CoNLL04 and SciERC development set) increases significantly with the number of negative samples.}
    \label{fig:neg_sampling}
\end{figure}

We also study the effect of the number and sampling of negative training examples. Figure \ref{fig:neg_sampling} shows the F1 score (relations and entities) for the CoNLL04 and SciERC development sets, plotted against the number of negative samples $N_e/N_r$ per sentence. We see that a sufficient number of negative samples is essential: When using only a single negative entity and relation ($N_e{=}N_r{=}1$) per sentence, relation F1 is about $10.5$\% (CoNLL04) and $9.7$\% (SciERC). 
With a high number of negative samples, the performance stagnates for both datasets. However, we found our results to be more stable when using a sufficiently high $N_e$ and $N_r$ (we chose $N_e{=}N_r{=}100$ in all other experiments).

For relation classification, we also assess the effect of using weak instead of strong negative relation samples: Instead of using the entity classifier as a filter for entity candidates $\mathcal{S}$ and drawing {\it strong} negative training samples from $\mathcal{S} {\times} \mathcal{S}$, we omit span filtering and sample random training span pairs not matching any ground truth relation. 
With these {\it weak} samples, our model retains a high recall ($84.4$\%) on the CoNLL04 development set, but the precision decreases drastically to about $4.3$\%. We observed that the model tends to predict {\it subspans} of entities to be in relation when using weak samples: For example, in the sentence \enquote{[John Wilkes Booth]$_{\text{head}}$, who assassinated [President Lincoln]$_{\text{tail}}$, was an actor}, the pairs (\enquote{John}, \enquote{President}) or (\enquote{Wilkes}, \enquote{Lincoln}) are chosen. Additionally, pairs where one entity is correct and the other one incorrect are also favored by the model. Obviously, span filtering is not only beneficial in terms of training and evaluation speed, but is also vital for accurate localization in SpERT.

\subsection{Localized context}

Despite advances in detecting long distance relations using LSTMs or the attention mechanism, the noise induced with increasing context remains a challenge. By using a {\it localized} context, i.e. the context between entity candidates, the relation classifier can focus on the sentence's section that is often most discriminative for the relation type. To assess this effect, we compare localized context with two other context representations that use the whole sentence:

\begin{itemize}
    \item \textbf{Full context}: Instead of performing a max pooling over the context between entity candidates, a max pooling over all tokens in the sentence is conducted.
    \item \textbf{Cls token}: Just like in the {\it entity} classifier (Figure \ref{fig:approach}, green), we use a special classifier token as context, which is able to attend to the whole sentence.
\end{itemize}

\begin{figure}[ht!]
    \centering
    \includegraphics[width=0.45\textwidth]{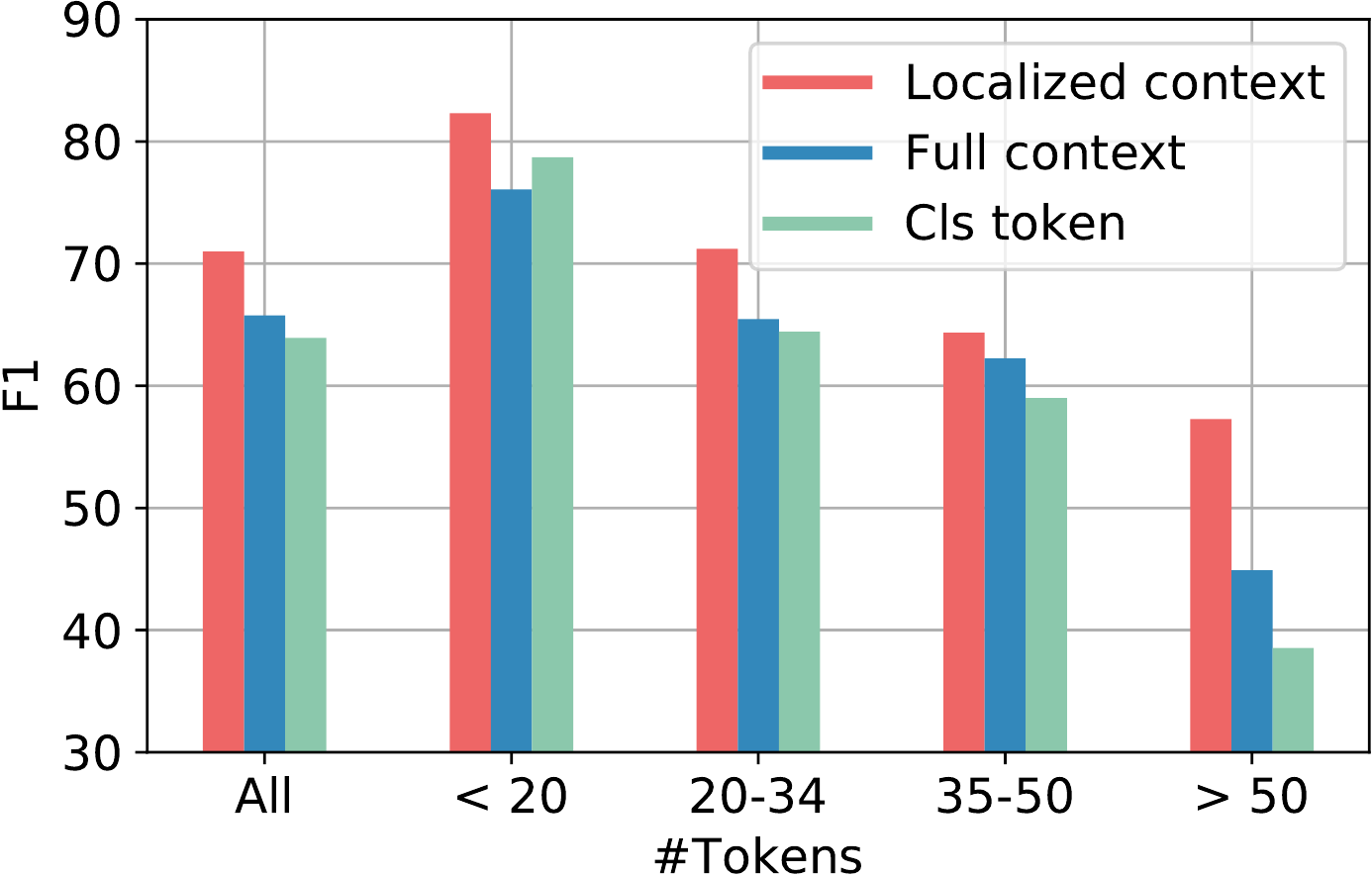}
    \caption{Macro F1 scores of relation classification on the CoNLL04 development set when using different context representations. Localized context (red) performs best overall (left), particularly on long sentences with ${>}50$ tokens (right).
}
        \label{fig:localized_context}
\end{figure}

We evaluate the three options on the CoNLL04 development set (Figure \ref{fig:localized_context}): When employing SpERT with a localized context, the model reaches an F1 score of $71.0$\%, which significantly outperforms a max pooling over the whole sentence ($65.8$\%) and using the classifier token ($63.9$\%).

Figure \ref{fig:localized_context} also displays results with respect to the sentence length: We split the CoNLL04 development set into four different parts, namely sentences with ${<}20, 20-34, 35-50$ and ${>}50$ tokens. Obviously, localized context leads to comparable or better results for all sentence lengths, particularly for very long sentences: Here, it reaches an F1 score of $57.3$\%, while the performance drastically decreases to $44.9/38.5$\% when using the other options. Table \ref{table:examples} (middle) shows an example of a long sentence with multiple entities: By using a localized context the model correctly predicts the three \emph{Located-In} relations, while relying on the full context leads to many false positive relations such as (\enquote{Jackson}, \emph{Located-In}, \enquote{Colo.}) or (\enquote{Wyo.}, \emph{Located-In}, \enquote{McAllen}).
This shows that guiding the model towards relevant sections of the input sentence is vital. An interesting direction for future work is to learn the relevant context with respect to the entity candidates, and to incorporate precomputed syntactical information into SpERT.

\subsection{Pre-training and entity representation}

Next, we assess the effect of BERT's language modeling pre-training. It seems intuitive that pre-training on large-scale datasets helps the model to learn semantic and syntactic relations that are hard to capture on a limited-scale target dataset. Therefore, we test three variants of pre-training:
\begin{enumerate}
    \item {\bf Full}: We use the fully pre-trained BERT model (\emph{LM Pre-trained}, our default setting).
    \item {\bf --Layers}: We retain pre-trained token embeddings but train the layers from scratch (using the default initalization~\cite{devlin:2018:bert}).
    \item {\bf --Layers,Embeddings}: We train layers and token embeddings from scratch (again, using the default initialization).
\end{enumerate}

As Table \ref{table:pretraining} shows, training the BERT layers from scratch results in a performance drop of about $17.0$\% and $29.4$\% (macro) F1 for entity and relation extraction respectively.

Further, training the token embeddings from scratch results in an even stronger drop in F1. These results suggest that pre-training a large network like BERT is challenging on the fairly small joint entity and relation extraction datasets. Therefore, language modeling pre-training is vital for generalization and to obtain a competitive performance.

Finally, we investigate different options for the entity span representation $\mathbf{e}(s)$ other than conducting a max pooling over the entity's tokens, namely a sum and average pooling (note that a size embedding and a context representation is again concatenated to obtain the final entity representation (Equation \ref{eq:entityrepr})).
Table \ref{table:architecture} shows the CoNLL04 (macro) F1 with respect to the different entity representations: We found the averaging of the entity tokens to be unsuitable for both entity ($69.2$\%) and relation extraction ($44.8$\%). Sum pooling improves the performance to $80.3/68.2$\%. Max pooling, however, outperforms this by another increase of $3.8$\% and $2.8$\% respectively.

\begin{table}
\sisetup{round-mode=places,detect-weight}
\centering
\begin{tabular}{l S S}
\toprule
    \textbf{Pre-training} & \textbf{Entity F1} & \textbf{Relation F1} \\ \midrule
    Full & 84.0382782807985 & 70.98481070981288 \\
    -- Layers & 67.06030686520602 & 41.57574335102769 \\
    -- Layers,Embeddings & 50.84065134898433 & 25.220059214696562 \\
\bottomrule
\end{tabular}
\caption{Effect of BERT pre-training on entity and relation extraction (CoNLL04 development set). A fully pre-trained BERT model significantly outperforms two BERTs in which the self-attention layers (--Layers) or the layers and the BPE input token embeddings (--Layers,Embeddings) are trained from scratch.}
\label{table:pretraining} 
\end{table}
\begin{table}
\sisetup{round-mode=places,detect-weight}
\centering
\begin{tabular}{l S S}
\toprule
    \textbf{Pooling} & \textbf{Entity F1} & \textbf{Relation F1} \\ \midrule
    
    Max & 84.0382782807985 & 70.98481070981288 \\
    Sum & 80.25535077591576 & 68.16252726255951 \\
    Average & 69.20729693966761 & 44.746487372845266 \\
\bottomrule
\end{tabular}
\caption{Investigation of different entity span representations $\mathbf{e}(s)$ (summing and averaging of entity's tokens) on the CoNLL04 development set.}
\label{table:architecture} 
\end{table}

\begin{table*}
\centering
\begin{tabular}{l p{14.4cm}}
\toprule
    \multicolumn{2}{l}{{\bf (a) Examples of Overlapping Entities}} \\ \midrule
     & Six days after starting acyclovir she exhibited signs of \gb{\gb{lithium} toxicity}.  \\ \midrule
     & A diagnosis of masked \gb{\gb{theophylline} poisoning} should be considered in similar situations involving a rapid decrease of insulin requirements. \\
    \midrule
    \midrule
    \multicolumn{2}{l}{{\bf (b) Effect of Localized Context}} \\ \midrule     localized context 
    &  Temperatures around the nation at 2 a.m. EST ranged from 2 degrees at \gb{Jackson}$_1$, \gb{Wyo.}$_1$, and \gb{Gunnison}$_2$, \gb{Colo.}$_2$, to 89 degrees at \gb{McAllen}$_3$, \gb{Texas}$_3$. \\ \midrule
     full context &  Temperatures around the nation at 2 a.m. EST ranged from 2 degrees at \bb{\bb{Jackson}$_1$}$_2$, \bb{Wyo.}$_3$, and \gb{Gunnison}$_4$, \bb{\gb{Colo.}$_4$}$_1$, to 89 degrees at \bb{\gb{McAllen}$_5$}$_3$, \bb{\gb{Texas}$_5$}$_2$. \\
    \midrule
    \midrule
    \multicolumn{2}{l}{{\bf (c) Error Cases}} \\ \midrule           incorrect spans & \bb{Delayed \rb{bowel injury}} is an infrequently observed complication of \bb{\rb{chromic phosphate}} administration.\\ \midrule
           
       
       syntax & Ambassador Miller is also scheduled to meet with Crimean Deputy \bb{Yevhen Saburov} and \bb{\gb{Black Sea Fleet}} Commander \gb{Eduard Baltin}.\\ \midrule
       
      logical &  \rb{Becton Dickinson} sells needle containers to doctors and hospitals but may develop a container for home use, said \rb{Linda Schmitt}, an assistant product manager.\\ \midrule
       
      classification & Finally, we briefly describe an experiment which we have done in extending the  \rb{\bb{n-best speech / language integration architecture}$_{\text{rel:Used-For}}$}$_{\text{rel:Evaluate-For}}$ to improving \rb{\bb{OCR accuracy}$_{\text{rel:Used-For}}$}$_{\text{rel:Evaluate-For}}$. \\ \midrule
           missing annotation & \bb{\gb{Norton Winfred Simon}} was born on Feb. 5, 1907, in \gb{Portland, Ore.}, and spent his teenage years in \bb{San Francisco} .\\
     \bottomrule
\end{tabular}
\caption[]{SpERT relation extraction examples showing that (a) as a span-based approach, our model can deal with overlapping entities, and (b) localized context yields better precision for long sentences compared to using the full sentence as context. (c) showcases various common sources of error.
\textcolor{darkgreen}{green [*]} = true positive relation, \textcolor{blue}{blue [*]} = false positive relation, \textcolor{red}{red [*]} = false negative relation.} 
\label{table:examples} 
\end{table*}

\subsection{Error inspection}
Although SpERT yields strong results on joint entity and relation extraction, we observed several common errors which leave room for further research. Table \ref{table:examples} (bottom) contains examples of five error cases we found to be common in the evaluated datasets:

\begin{itemize}
    \item \textbf{Incorrect spans}: One common source of error is the prediction of slightly incorrect entity spans, e.g. by adding a nearby word or missing a word annotated in the ground truth. This error occurs especially often in the domain specific ADE and SciERC datasets. 
    \item \textbf{Syntax}: Another frequently encountered error is the prediction of a relation (here: \emph{Work-For}) between two entities, which could possibly be related based on their entity types (\enquote{Yevhen  Saburov}, a person, and \enquote{Black Sea Fleet}, an employer), but are not related in the sentence.
    \item \textbf{Logical}: Sometimes, a relation is not explicitly stated in the sentence, but can logically be inferred based on the context. In the depicted case, it is not stated that \enquote{Linda Schmitt} is indeed a product manager of \enquote{Becton Dickinson}, but it is obvious due to her speaking for the company.
    \item \textbf{Classification}: In some rare cases (especially in the SciERC dataset), SpERT correctly predicted two related entities, but assigned a wrong relation type. 
    \item \textbf{Missing annotation}: Finally, there are also some cases where a correct prediction is missing in the ground truth. Here, in addition to correctly predicting (\enquote{Norton Winfried Simon}, \emph{Live-In}, \enquote{Portland, Ore.}), SpERT also outputs (\enquote{Norton Winfried Simon}, \emph{Live-In}, \enquote{San Francisco}), which is correct but not labeled.
\end{itemize}

\section{CONCLUSIONS}
We have presented SpERT, a span-based model for joint entity and relation extraction that relies on the pre-trained Transformer network BERT as its core. We show that with strong negative sampling, span filtering, and a localized context representation, a search over all spans in an input sentence becomes feasible. Our results suggest that span-based approaches perform competitive to BILOU-based models and may be the more promising approach for future research due to their ability to identify overlapping entities.

In the future, we plan to investigate more elaborate forms of context for relation classifiers. Currently, our model simply employs the span between the two entities, which proved superior to the full context. Employing additional syntactic features or learned context -- while maintaining an efficient exhaustive search -- appears to be a promising challenge.

\ack This work was funded by German Federal Ministry of Education and Research (Program FHprofUnt, Project DeepCA (13FH011PX6)).

\bibliography{spert}
\end{document}